\pdfoutput=1
\documentclass[11pt]{article}

\usepackage{EMNLP2023}

\usepackage{times}
\usepackage{latexsym}
\usepackage{graphicx}
\usepackage{url}
\usepackage[T1]{fontenc}
\usepackage[utf8]{inputenc}

\usepackage{microtype}

\usepackage{inconsolata}

\title{Evaluating the Logical Reasoning Ability of ChatGPT and GPT-4 }

\author{Hanmeng Liu \\
  Westlake University \\
  \texttt{liuhanmeng@westlake.edu.cn} \\\And
  Ruoxi Ning \\
  Zhejiang University \\
  \texttt{ruoxining@zju.edu.cn} \\\And
  Zhiyang Teng \\
  Nanyang Technological University \\
  \texttt{zhiyang.teng@ntu.edu.sg} \\\AND
  Jian Liu \\
  Fudan University \\
  \texttt{jianliu17@fudan.edu.cn} \\\And
  Qiji Zhou \and  Yue Zhang \thanks{Yue Zhang is the corresponding author}\\
  Westlake University \\
  \texttt{{zhouqiji, zhangyue}@westlake.edu.cn} \\
  }

\begin{document}
\maketitle
\begin{abstract}
 Harnessing logical reasoning ability is a comprehensive natural language understanding endeavor. With the release of Generative Pretrained Transformer 4 (GPT-4), highlighted as "advanced" at reasoning tasks, we are eager to learn the GPT-4 performance on various logical reasoning tasks. This report analyses multiple logical reasoning datasets, with popular benchmarks like LogiQA and ReClor, and newly-released datasets like AR-LSAT. We test the multi-choice reading comprehension and natural language inference tasks with benchmarks requiring logical reasoning. We further construct a logical reasoning out-of-distribution dataset to investigate the robustness of ChatGPT and GPT-4. We also make a performance comparison between ChatGPT and GPT-4. Experiment results show that ChatGPT performs significantly better than the RoBERTa fine-tuning method on most logical reasoning benchmarks. With early access to the GPT-4 API we are able to conduct intense experiments on the GPT-4 model. The results show GPT-4 yields even higher performance on most logical reasoning datasets. Among benchmarks, ChatGPT and GPT-4 do relatively well on well-known datasets like LogiQA and ReClor. However, the performance drops significantly when handling newly released and out-of-distribution datasets. Logical reasoning remains challenging for ChatGPT and GPT-4, especially on out-of-distribution and natural language inference datasets. We release the prompt-style logical reasoning datasets as a benchmark suite and name it \textbf{LogiEval}.
\end{abstract}

\section{Introduction}
Logical reasoning is essential to human intelligence, and incorporating logical reasoning abilities into natural language understanding (NLU) systems has been an active research interest from the beginning of artificial intelligence \cite{cresswell1973logics} \cite{kowalski1979logic} \cite{iwanska1993logical}.
Researchers have been exploring various approaches to achieve this goal, including rule-based methods, symbolic systems \cite{maccartney-manning-2007-natural}, fine-tuning large language models \cite{wang-etal-2018-glue}, and combining both neural and symbolic approaches \cite{li-srikumar-2019-augmenting}.

In the traditional logical and semantic approach, computational linguists developed symbolic systems utilizing First-Order-Logic (FOL) or Natural Logic \cite{maccartney-manning-2007-natural} to tackle fundamental inference tasks. Rule-based models struggle to unravel problems like the RTE challenge \cite{Dagan} with hand-crafted rules and theorem provers. 
Formal logic reasoning adopted by early researchers came up with symbolic systems and hand-crafted rules, where knowledge was represented explicitly using formal logic or other symbolic representations. With rules, the systems can process deduction operations. However, these approaches face challenges in handling ambiguity and scalability. They are brittle when dealing with real-world natural language data.

%With the advent of deep learning era, especially the advance of Transformer-based architecture, NLU research has been push forward significantly \cite{Devlin} \cite{lewis2019bart}, with competitive results being reported over several tasks, including natural language inference (NLI) \cite{Dagan} \cite{maccartney-manning-2007-natural} and machine reading comprehension (MRC) \cite{hermann2015teaching} \cite{sar}. 

\begin{figure}[t!]
\centering
\setlength{\abovecaptionskip}{0.1cm}
\setlength{\belowcaptionskip}{0.2cm}
\includegraphics[width=0.48\textwidth]{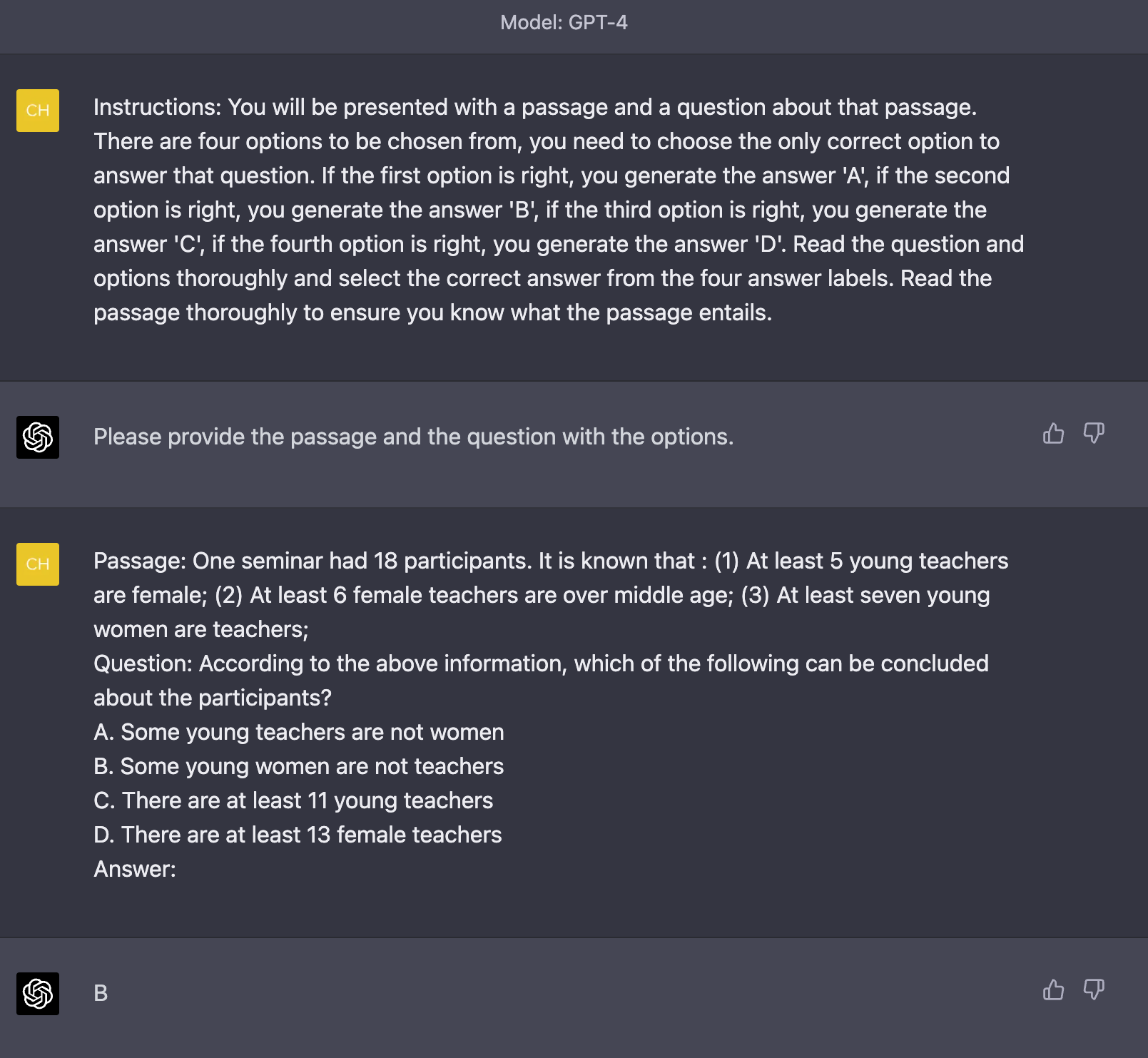}
\caption{Instruction and question format for logical reading comprehension tasks.  }
\label{fig:showcase}
%\vspace{-0.5cm}
\end{figure}

\begin{figure}[t!]
\centering
\setlength{\abovecaptionskip}{0.1cm}
\setlength{\belowcaptionskip}{0.2cm}
\includegraphics[width=0.48\textwidth]{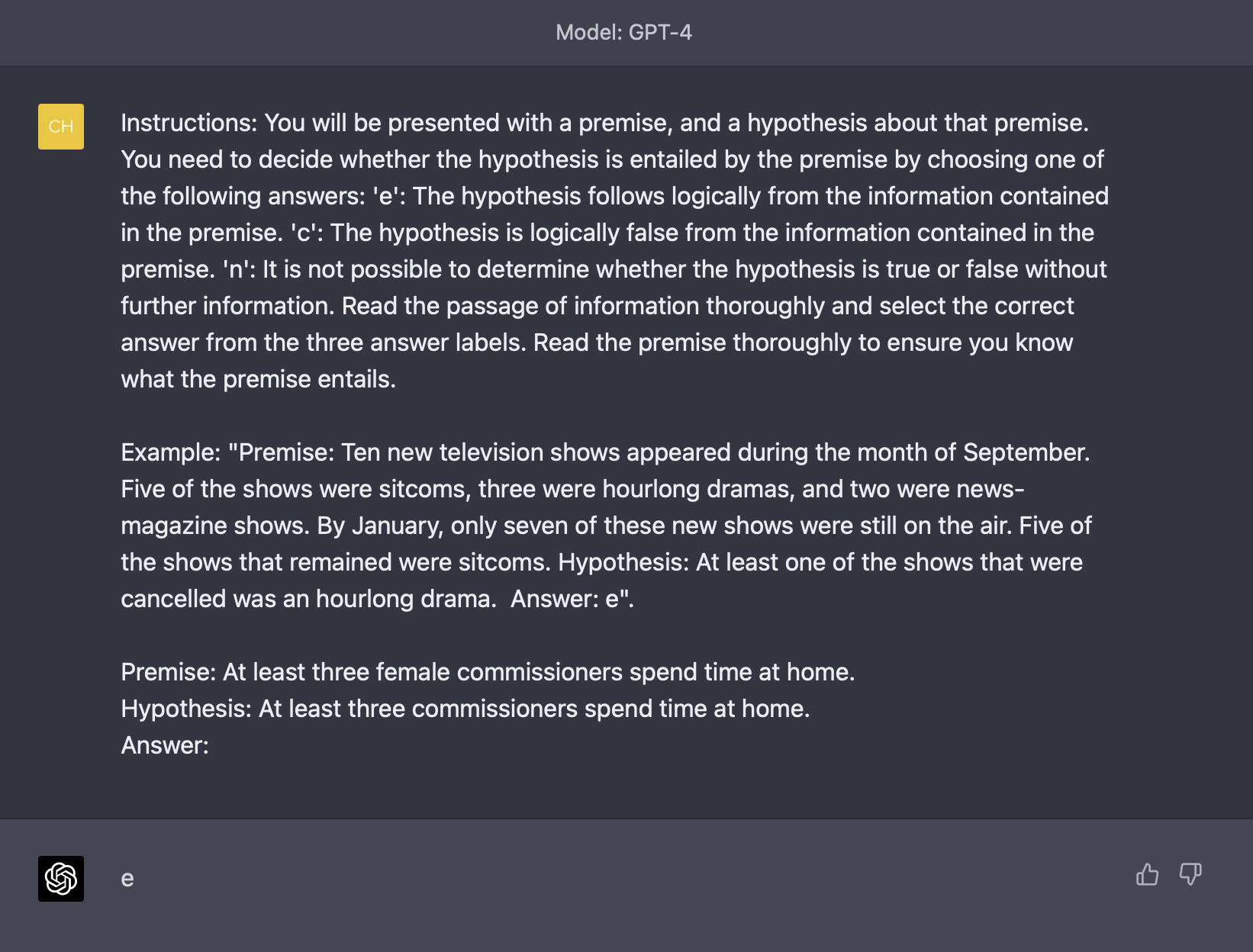}
\caption{Instruction and question format for natural language inference tasks.  }
\label{fig:showcasenli}
%\vspace{-0.5cm}
\end{figure} 

The era of neural network models sees the rise of large-scale NLI datasets as popular benchmarks. For example, the SNLI \cite{bowman-etal-2015-large} and the Multi-genre NLI (MNLI) \cite{N18-1101} datasets are created through crowd-sourcing, featuring an immense data size and broad coverage. They catalyze the development of models with better representation abilities and become the go-to benchmark for natural language understanding research. The giant leap in model performance comes with the advent of Transformer-based \cite{DBLP:journals/corr/VaswaniSPUJGKP17} language models like BERT \cite{DBLP:journals/corr/abs-1810-04805} when the training schemes of such models enable them to access colossal unlabelled corpora. As a result, building language models with trillions of parameters become possible \cite{DBLP:journals/corr/abs-2005-14165} \cite{DBLP:journals/corr/abs-1910-10683}. The paradigm of pre-training and fine-tuning has since become the dominant solution to textual inference tasks. Researchers fine-tune language models on task-specific datasets after pre-training models on massive text corpora. Large pre-trained language models (LMs) achieve beyond-human performances on popular NLI and MRC benchmarks, urging for more sophisticated benchmarks in textual inference.

NLP research on logical reasoning regains momentum with the recent releases of a handful of datasets, notably LogiQA and Reclor. The datasets are collected from logical reasoning exams such as Chinese Civil Servant Exams and Law School Admission Test (LSAT). These tests are challenging even for humans and are golden-labeled data with good quality. Logical reasoning is exploited in many probing tasks over large Pre-trained Language Models (PLMs) and downstream tasks like question-answering and dialogue systems. PLMs perform poorly compared to traditional benchmarks. Despite the progress made so far, achieving human-like logical reasoning capabilities in NLU systems remains a challenging task. 

Generative Pre-trained Transformer 4 (GPT-4) \cite{openai2023gpt4}, as well as ChatGPT, is a newly-released language model developed by OpenAI, designed to understand and generate multi-modal contents. GPT-4 is promoted to possess even more powerful capabilities in tasks that require logical reasoning. Logical reasoning is essential to human intelligence, enabling us to draw conclusions, make predictions, and solve problems based on given information. Incorporating logical reasoning into language models like GPT-4 can revolutionize natural language understanding (NLU) systems, making them more accurate, robust, and capable of understanding complex information in natural language. 

The evaluation of ChatGPT and GPT-4 for logical reasoning tasks explores their performance on several logical reasoning benchmarks, detailing the strengths and limitations of ChatGPT and GPT-4 in this domain. Previous ChatGPT evaluation indicates that ChatGPT performs better on deductive reasoning than inductive ones \cite{bang2023multitask} and that ChatGPT performs better on factual inputs \cite{qin2023chatgpt}. After the release of GPT-4, we will discuss two tasks for evaluating its logical reasoning abilities: multi-choice reading comprehension and natural language inference. Both tasks are reasoning-heavy and serve as a playground for testing models' reasoning abilities. Several logical reasoning datasets have been released under the two tasks. These benchmarks proved to be difficult to solve for PLMs. We hope this report will shed more light on the logical reasoning ability of ChatGPT and GPT-4. Our contributions are as follows:

1. We test ChatGPT and GPT-4 on two logical reasoning tasks: multi-choice reading comprehension and natural language inference. We conduct experiments on multiple logical reasoning benchmarks to analyze the logical reasoning ability of ChatGPT and GPT-4.

2. We introduce LogiEval, a test suite for testing prompt-based large language models. We release the benchmark at \url{https://github.com/csitfun/LogiEval} 

3. Our experiments show that both ChatGPT and GPT-4 are good at solving well-known logical reasoning reading comprehension benchmarks but struggle at handling out-of-distribution datasets. Their performances on natural language inference tasks requiring logical reasoning still need improving.

\section{Evaluation Settings}
We consider multi-choice reading comprehension and natural language inference tasks for our evaluation. Multi-choice reading comprehension is heavily tested on large language models for these tasks often have clearly-formed and high-quality datasets. On the other hand, the natural language inference task is a fundamental task for evaluating reasoning abilities.
The datasets and the evaluation settings are provided as follows.
\subsection{Datasets}

\begin{figure}[t!]
\centering
\setlength{\abovecaptionskip}{0.1cm}
\setlength{\belowcaptionskip}{0.2cm}
\includegraphics[width=0.48\textwidth]{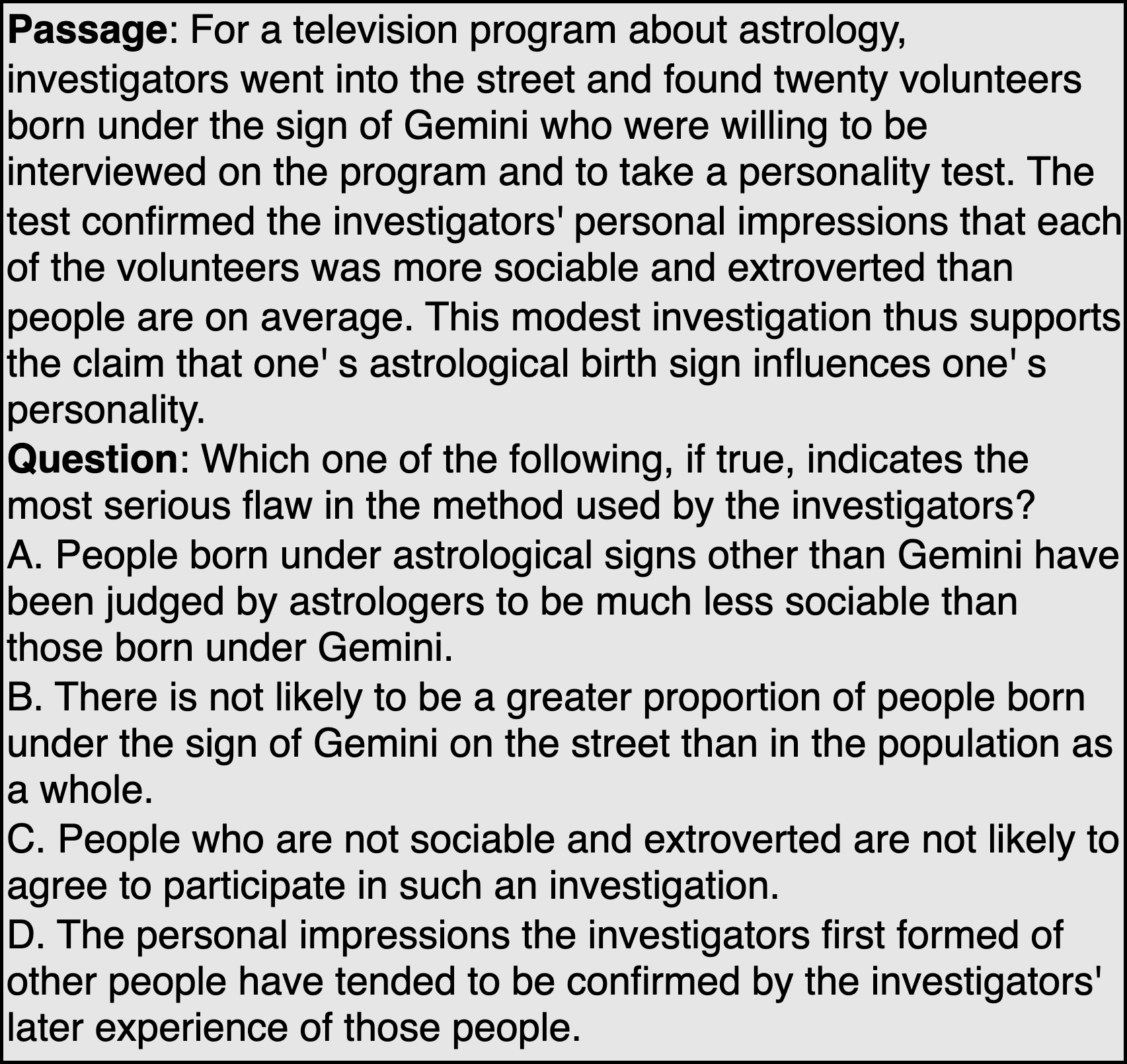}
\caption{A multi-choice reading comprehension example from the LogiQA 2.0 dataset.  }
\label{fig:logiqa}
%\vspace{-0.5cm}
\end{figure} 

\begin{figure}[t!]
\centering
\setlength{\abovecaptionskip}{0.1cm}
\setlength{\belowcaptionskip}{0.2cm}
\includegraphics[width=0.48\textwidth]{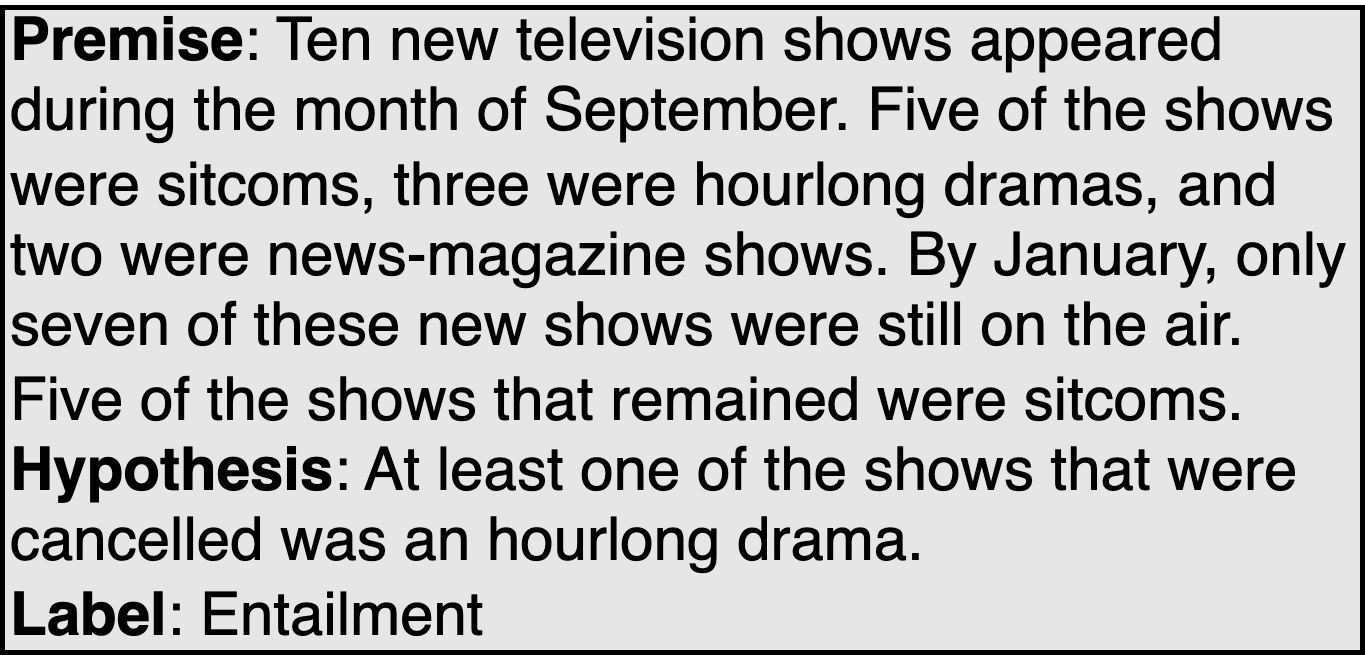}
\caption{An NLI example from the ConTRoL dataset.  }
\label{fig:control}
%\vspace{-0.5cm}
\end{figure}

\subsubsection{Multi-choice Reading Comprehension}
Machine reading is a popular task in NLP. In the typical multiple-choice task setting, given a passage and a question, a system is asked to select the most appropriate answer from a list of candidate answers.

\textbf{LogiQA} is a dataset specifically designed for multi-choice question-answering tasks that involve logical reasoning. The dataset is adopted from the Chinese Civil Service Examination, translated into English, and released in Chinese and English. The dataset has been updated to the 2.0 version \footnote{\url{https://github.com/csitfun/LogiQA2.0}}, where the data size has been enlarged. We choose the test sets of both the Chinese and English versions for our evaluation. Figure \ref{fig:logiqa} shows an example from the LogiQA 2.0 test set. 

\textbf{ReClor} is another logical reasoning dataset designed for reading comprehension tasks requiring logical reasoning. It collects question-answering examples from the LSAT exams \footnote{\url{https://www.lsac.org/lsat}}, which are targeted to testing human logical reasoning abilities. We use the development set for our testing because the test set does not include gold labels.

\subsubsection{Natural Language Inference}
Natural language inference is the task of deciding the logical relationship between a hypothesis and a premise. The typical scheme is a text classification where the model needs to choose one from three labels: entailment, contradiction, and neutral.
%RuleTaker \cite{DBLP:journals/corr/abs-2002-05867} is a synthetic dataset to examine the reasoning ability of transformers\cite{DBLP:journals/corr/VaswaniSPUJGKP17} over natural language rules. With input facts and input rules as context, the output is a binary true-or-false answer. Although originally designed for question-answering, the dataset can be easily converted into NLI-style.

\begin{table*}
\centering
\scalebox{0.85}{
\begin{tabular}{l|c|c|c|c|c}
\hline
\textbf{Dataset} & \textbf{LogiQA 2.0 test} & \textbf{LogiQA 2.0 zh test}& \textbf{ReClor dev} & \textbf{AR-LSAT test} & \textbf{LogiQA 2.0 ood} \\
\hline
\textbf{Size} & 1572 &1594 & 500 & 230 & 1354 \\
\hline
Human avg. & 86.00 & 88.00 & 63.00 & 56.00 & 83.00 \\
human ceiling & 95.00 & 96.00 & 100.00 & 91.00 & 99.00 \\
\hline
RoBERTa &   48.76 & 35.64  & 55.01 & 23.14 & 33.22  \\
ChatGPT (API) & 52.37 & 53.18 & 57.38 & 20.42 & 38.44  \\
\hline
GPT-4 (Chat UI) & 75.26(73/97) & 51.76 (44/85) & 92.00 (92/100) & 18.27 (19/104) & 48.21(54/112) \\
GPT-4 (API) & 72.25 & 70.56 & 87.20 & 33.48 & 58.49   \\
\hline
\end{tabular}
}
\caption{\label{Tab:mcrq}
ChatGPT and GPT-4 performance on the Logical multi-choice machine reading comprehension task (accuracy \%). ``LogiQA 2.0 zh test'' refers to the test set of the LogiQA 2.0 Chinese version. ``LogiQA 2.0 ood'' represents the out-of-distribution data of LogiQA 2.0.
}
\end{table*}

\textbf{ConTRoL} \cite{DBLP:journals/corr/abs-2011-04864} is an NLI dataset that further investigates contextual reasoning under the NLI framework. It has 36.2\% of premise-hypothesis pairs under the category of logical reasoning. Figure \ref{fig:control} shows an example from the ConTRoL dataset.

\textbf{MED} \cite{yanaka2019neural} and \textbf{HELP} \cite{yanaka-EtAl:2019:starsem} are two NLI datasets focusing on monotonicity reasoning, which is an essential concept in Natural Logic \cite{maccartney2007natural}. The datasets are generated through monotonicity rules and only investigate monotonicity-related inference specifically. Similar to the NLI section of our dataset, MED and HELP investigate basic logic phenomena in natural language, which is monotonicity in particular. For the HELP dataset, we use the train set for our evaluation.
Here is an example from the HELP dataset:

Premise: Tom said that neither parents had ever been to Boston.

Hypothesis: Tom said that neither one of his parents had ever been to Boston.

Label: Entailment

\textbf{ConjNLI} \cite{DBLP:journals/corr/abs-2010-10418} is a challenging stress test for NLI over conjunctive sentences, where the premise differs from the hypothesis by having conjuncts being removed, added, or replaced. Logical reasoning about conjunctions is heavily tested in ConjNLI. Premise-hypothesis pairs are created automatically by applying conjunct operations on collected conjunctive sentences. 
Here is an example from the ConjNLI dataset:

Premise: In Quebec, an allophone is a resident, usually an immigrant, whose mother tongue or home language is neither French nor English.

Hypothesis: In Quebec, an allophone is a resident, usually an immigrant, whose mother tongue or home language is not French.

Label: Entailment

\textbf{TaxiNLI} \cite{DBLP:journals/corr/abs-2009-14505} is an NLI dataset re-annotated on the MNLI \cite{williams-etal-2018-broad} dataset with fine-grained category labels. The annotation includes logical categories like connectives, mathematical, and deduction. Notice that TaxiNLI is a subset of the MNLI dataset, so we include the MNLI dataset for our comparison as a traditional NLI benchmark. 
Here is an example from the TaxiNLI dataset:

Premise: and that you're very much right but the jury may or may not see it that way so you get a little anticipate you know anxious there and go well you know.

Hypothesis: Even if you're correct, I think the jury would pick up on that.

Label: Contradiction

\begin{figure}[t!]
\centering
\setlength{\abovecaptionskip}{0.1cm}
\setlength{\belowcaptionskip}{0.2cm}
\includegraphics[width=0.48\textwidth]{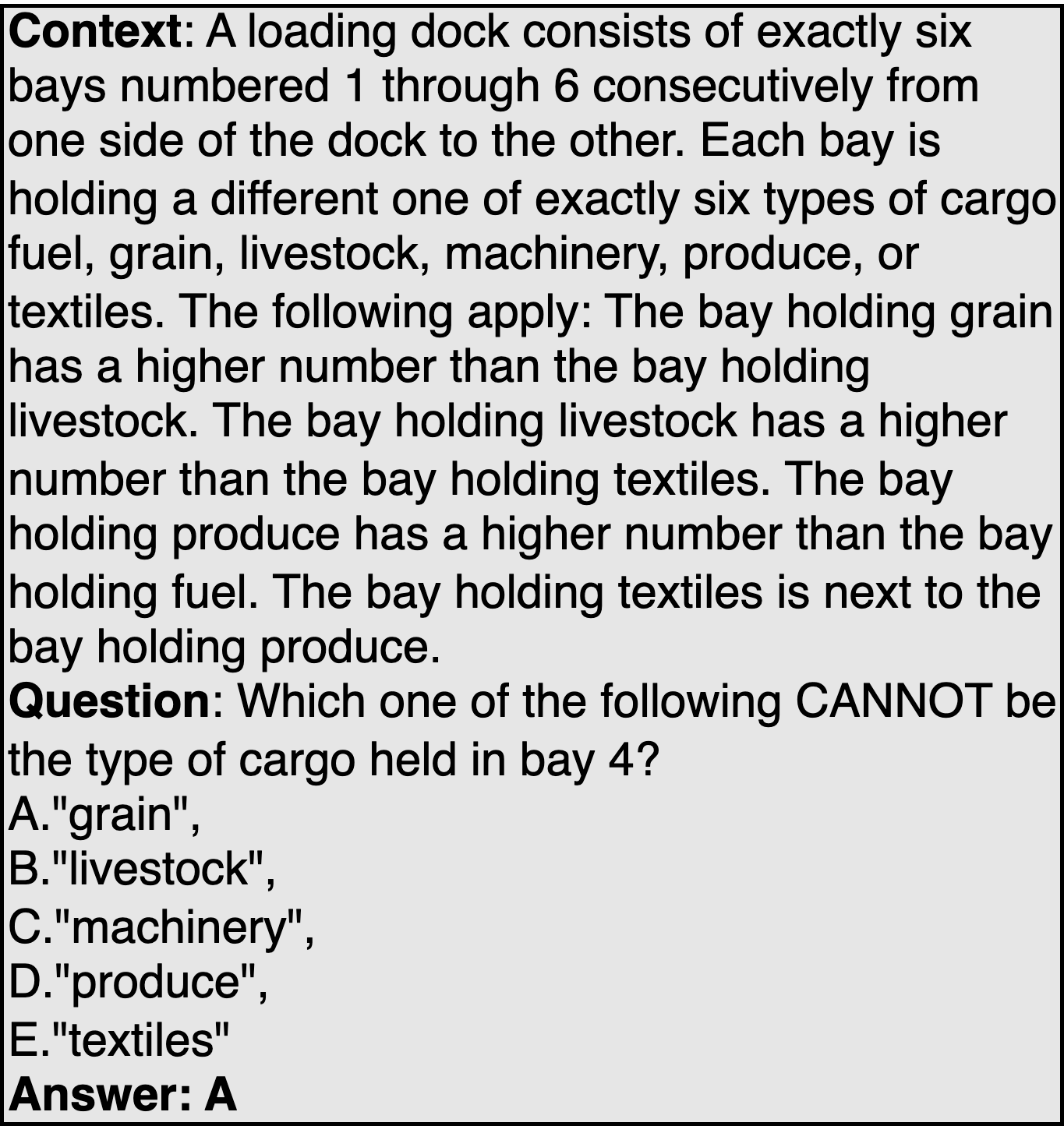}
\caption{An example from the AR-LSAT dataset.  }
\label{fig:arlsat}
%\vspace{-0.5cm}
\end{figure} 

\subsubsection{Out-of-distribution Data}
\textbf{AR-LSAT} \cite{wang2022lsat} is a new dataset of analytical reasoning questions from the Law School Admission Test. Released in 2022, it has 2064 questions, each describing a reasoning game belonging to three dominant types: (1) ordering game, (2) grouping game, and (3) assignment game. It is noticed that each question has five options rather than four. Figure \ref{fig:arlsat} shows an example from the AR-LSAT test set.

Besides, we construct a \textbf{LogiQA 2.0 out-of-distribution dataset}, which incorporates the newly released Chinese Civil Servant Exam from 2022. The test set is a collection of logical reasoning tests designed by experts from 2022 onwards. We release our data at \url{https://github.com/csitfun/LogiQA2.0}.

\begin{table*}
\centering
\scalebox{0.8}{
\begin{tabular}{l|c|c|c|c|c|c}
\hline
\textbf{Dataset} & \textbf{ConTRoL test} & \textbf{ConjNLI test}& \textbf{HELP} & \textbf{MED} & \textbf{TaxiNLI test} & \textbf{MNLI dev}\\
\hline
\textbf{Size} & 805 & 623 & 35891 & 5382 & 10071 & 9815 \\
\hline
Human avg. & 87.00 &   89.00  & 81.00 & 91.00 & 97.00 & 98.00 \\
Human ceiling & 94.00 &  100.00 &  95.00 & 99.00 & 100.00 & 100.00 \\
\hline
RoBERTa &   48.76 & 38.94 & 39.47 &   46.83 &  49.91   & 90.02  \\
ChatGPT (API) & 58.45 & 47.03 & 42.13 & 55.02 &  57.30  & 55.40\\
\hline
GPT-4 (Chat UI) & 58.18(64/110) & 61.00 (61/100) & 53.33 (56/105) & 75.79 (72/95) & 75.47(80/106) & 68.00 (68/100) \\
GPT-4 (API) & 56.40 & 72.71 & 46.01 & 89.42 & 60.08 & 64.08 \\
\hline
\end{tabular}
}
\caption{\label{Tab:nli}
ChatGPT and GPT-4 performance on the natural language inference task (accuracy \%).
}
\end{table*}

\subsection{Experiment Setting}

We take RoBERTa-base \cite{liu2019roberta} as our baseline model. Following a fine-tuning scheme, we use Huggingface's RoBERTa-base model as our pre-trained language model. RoBERTa-base is trained on the training set for 5 epochs for each dataset. We also set up a human baseline by reporting the average and ceiling performance of human testees.

For ChatGPT and GPT-4, we follow an instruction-prompt scheme for both Natural Language Inference and multi-choice reading comprehension tasks. Figure \ref{fig:showcase} shows the instruction for multi-choice reading comprehension tasks. 

After investigating the methods of prompt-designing for reasoning tasks, we find that there are mainly three types of prompt designing for NLI tasks, specifying the labels (entailment, neutral or contradiction)\cite{qin2023chatgpt}, specifying the reasoning method (induction or deduction)\cite{bang2023multitask}, and chain-of-thought reasoning\cite{kojima2023large} which will be specified in the next chapter. Among these, specifying the labels method suits our purpose for most of our NLI datasets are 3-label classification tasks. Thus, we prompt GPT with the 3 three possible relationships between the hypothesis and conclusion, entailment, contradiction and neutral, every time we ask a question to GPT. The instruction we use for the multi-choice reading comprehension task is in Appendix \ref{sec:appendix_a}.

Figure \ref{fig:showcasenli} shows the instruction for multi-choice reading comprehension tasks. 
The instruction we use for the natural language inference task is in Appendix \ref{sec:appendix_b}.

For evaluating ChatGPT, We use the Eval\footnote{\url{https://github.com/openai/evals}} framework provided by OpenAI, a suite for evaluating OpenAI models and an open-source registry of benchmarks.
The model we choose is ``gpt-3.5-turbo'' (version March 23, 2023). Apart from task structure, we offer an in-context example to each API call to guarantee controlled output.

GPT-4 has been limited access to subscribe users from March 14, 2023. We are granted early access to GPT-4 API by merging requests to the OpenAI Eval repository. So we are able to use the GPT-4 API and the OpenAI Eval framework. The model we use is ``default-gpt-4'' (version March 14, 2023). We also utilize the GPT-4 Chat UI to conduct our GPT-4 experiments and further analyses with two OpenAI Plus accounts.

\section{Results}
\subsection{Experiment results on the multi-choice reading comprehension tasks}
Table \ref{Tab:mcrq} shows the results of the multi-choice reading comprehension datasets. 
\subsubsection{The performance of ChatGPT}
ChatGPT shows a performance increase compared to the baseline model on several long-standing benchmarks. The accuracy of the LogiQA 2.0 test set is 53.37\%, nearly 4 points higher than the RoBERTa base model. The performance gap between ChatGPT and RoBERTa is salient when testing on the Chinese version of LogiQA 2.0, which indicates the performance consistency of ChatGPT in both Chinese and English languages. ChatGPT yields the best performance on the ReClor dataset with an accuracy of 57.38\%, compared with RoBERTa's 55.01\% accuracy. 
However, ChatGPT experiences a huge performance drop on out-of-distribution datasets. On the AR-LSAT test set, the accuracy is only 20.42\%, lower than the performance of RoBERTa base. On LogiQA 2.0 ood, the performance is 38.44\%, still lower than that of RoBERTa base. 
From the experiments above, ChatGPT performs well on well-known Logical reasoning like LogiQA and ReClor. The accuracy of ChatGPT surpasses fine-tuning methods by a small margin. However, when tested on the newly released dataset, namely AR-LSAT, and on LogiQA 2.0 out-of-distribution dataset, the performance declined significantly.

Despite its limitations, ChatGPT still represents a significant advancement in natural language understanding and demonstrates the potential of language models to reason logically.

\subsubsection{The performance of GPT-4}
GPT-4 performs remarkably better than ChatGPT when doing manual tests on LogiQA and ReClor. On the LogiQA 2.0 test set (1572 instances), GPT-4 yields an accuracy of 72.25\%. On the Chinese version of the LogiQA 2.0 test set (1594 instances), the accuracy is 70.56, which is similar to the performance on the English version. On the ReClor dev set (500 instances, ReClor does not include gold labels on its test), GPT-4 reaches an 87.20\% accuracy which is the highest score among all three models. However, when tested on the AR-LSAT test set (230 intances), GPT-4 performs surprisingly worse with only a 33.48\% accuracy. The test result on LogiQA 2.0 ood data (1354 instances) shows that GPT-4 gets 58.49\% correctness, which is significantly lower than that on the LogiQA 2.0 test set. Nevertheless, the performance is still the highest among all three models. We will not haste to the conclusion, but it's safe to say that GPT-4's performance drop on out-of-distribution datasets is noticeable. For comparison, the GPT-4 Chat UI results are also provided, where we manually test a handful of data instances.

\subsection{Experiment results on the natural language inference task}
Table \ref{Tab:nli} shows the results on the natural language inference datasets.

\subsubsection{The performance of ChatGPT}
ChatGPT performs better than the RoBERTa model on the logical reasoning NLI datasets we test. On the ConTRoL test set, the accuracy is 58.45\%, higher than the RoBERTa-base model by nearly 10 percent. On the ConjNLI test set, ChatGPT yields 47.03\% accuracy, which outperforms RoBERTa by around 9 percent. On the HELP dataset, ChatGPT gets a 42.31\% accuracy, around 3 points higher than that of RoBERTa. On the MED dataset, ChatGPT gives 55.02\% accuracy, nearly 9 percent higher than that of RoBERTa. On the TaxiNLI test set, ChatGPT gives 57.30\% accuracy, over 7 percent higher than that of RoBERTa. For comparison, ChatGPT gives 55.40\% accuracy on the MNLI dev set, which is significantly lower than that of RoBERTa, which indicates that ChatGPT is not optimized for answering three-labeled natural language inference questions. Since it is noticed that ChatGPT is not good at following NLI task instructions, we provide an in-context example to help the model generate task labels, as shown in Figure \ref{fig:showcasenli}.
Overall, the results show that ChatGPT surpasses fine-tuned RoBERTa by only a small margin for logical reasoning NLI datasets. 

\subsubsection{The performance of GPT-4}
We test GPT-4's performance on logical reasoning NLI datasets. On the ConTRoL test set (805 instances), GPT-4 performs slightly lower than ChatGPT, yielding a 56.40\% accuracy. The performance of GPT-4 on the ConjNLI test (623 instances) and the MED test (5382 instances) is significantly better, with an accuracy of 72.71\% and 89.42\%, respectively. However, on the HELP (35891 instances) and TaxiNLI test (10071 instances), the performance of GPT-4 is slightly better than that of RoBERTa and ChatGPT, with an accuracy of 46.01\% and 60.08\%, respectively. The GPT-4 performance on the MNLI dev set is slightly better than on the TaxiNLI test, and yields 64.08\% accuracy, which indicates logical reasoning adds more challenges to the GPT-4 model. We also include the testing results with GPT-4 Chat UI and around 100 data instances for each NLI dataset.

The results on the six NLI datasets indicate that GPT-4 does not perform highly on logical reasoning natural language inference compared to multi-choice reading comprehension. We also notice that GPT-4 cannot output labels steadily even though the instruction is provided in the natural language inference task scenario. From this, we infer that GPT-4 is not good at following the instruction for the natural language inference task, though it is well-trained to follow the instruction for the multi-choice reading comprehension task.

\begin{figure}[t!]
\centering
\setlength{\abovecaptionskip}{0.1cm}
\setlength{\belowcaptionskip}{0.2cm}
\includegraphics[width=0.48\textwidth]{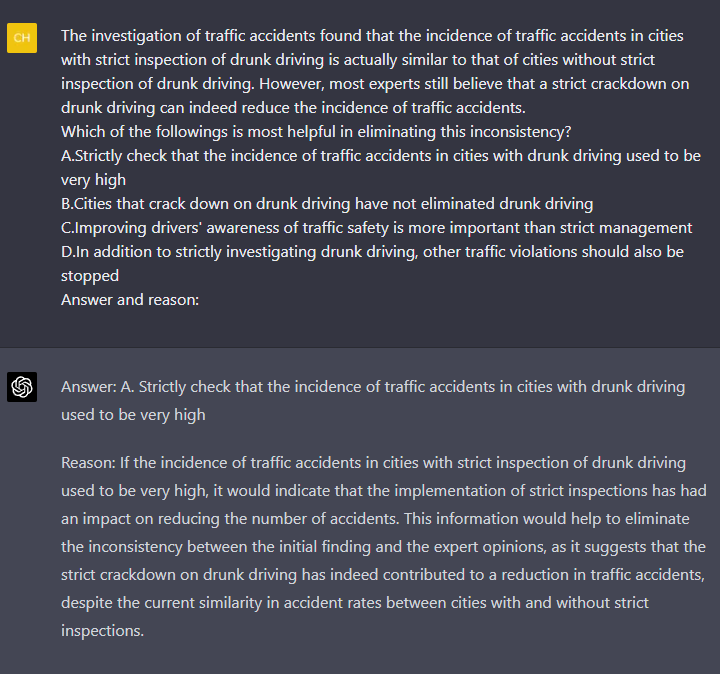}
\caption{An GPT-4 generation example where the model answers the question correctly.  }
\label{fig:reason}
%\vspace{-0.5cm}
\end{figure} 

\begin{figure}[t!]
\centering
\setlength{\abovecaptionskip}{0.1cm}
\setlength{\belowcaptionskip}{0.2cm}
\includegraphics[width=0.48\textwidth]{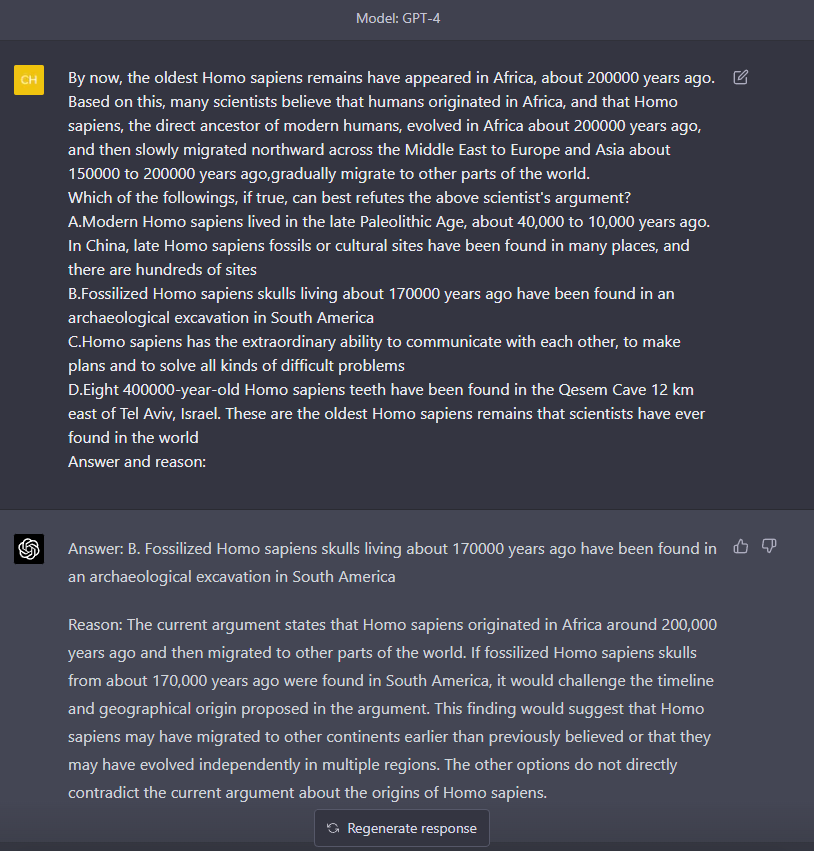}
\caption{An GPT-4 generation example where the model answers the question incorrectly.   }
\label{fig:reason_wrong}
%\vspace{-0.5cm}
\end{figure} 

\section{Analysis}
The experiment results show that ChatGPT and GPT-4 surpass RoBERTa on most logical reasoning benchmarks, including popular benchmarks like LogiQA and ReClor and less-known datasets like AR-LSAT. However, the performance drop on out-of-distribution datasets is noticeable for both GPT models, indicating they struggle to handle new and unfamiliar data. Thus, we conduct more case studies with the GPT-4 chat UI and further analyze its abilities.

\subsection{Answer and Reason}
For GPT-4 manual tests, we record the answers GPT-4 gives and the reasoning for the answer. Figure \ref{fig:reason} gives an example of GPT-4's answer and reasoning. In this example, GPT-4 did it correctly. From the paragraph's inconsistency between the similar drunk driving rate both with and without drunk driving checks, and the claim that the strict checks lower the drunk driving rate, GPT-4 chooses a fact prior to the contemporary situation that the drunk driving rate used to be high before strict checks, to solve this contradiction.

Figure \ref{fig:reason_wrong} shows an example where GPT-4 generates the wrong answer and reasoning. The passage talks about whether humans originated from Africa. The conclusion would be challenged if we find older fossils in other parts of the world. GPT-4 does not have a thorough understanding of the problem. It only focuses on establishing the chronological order of events. The failure to attend to the location of the fossils leads to the wrong answer.

In our assessment of GPT-4 on the logiQA dataset, we inspected the first 10 errors made by the model. Four were categorized as logical errors, such as affirming the question, negating the hypothesis, and begging the question. Three were identified as scope errors, including attributing predicates to incorrect subjects or assigning characters to the wrong objects. The remaining three errors fall in the inability to resolve semantic ambiguity, wherein GPT-4 selected a merely qualified response when asked for an optimal one. However, this limited sample of errors does not entail the conclusion that GPT-4 is incompetent to handle reasoning questions, as there are also instances where it accurately identifies the correct answer. GPT-4's occasional selection of wrong answers suggests that further examination is necessary to determine whether some features in the questions may trigger such incorrect choices.

%to do: explain more about the example

\begin{table}
\centering
\begin{tabular}{l|c|c}
\hline
\textbf{Dataset} & \textbf{LogiQA 2.0 ood} & \textbf{ConTRoL} \\
\hline
\# instances & 20 & 20 \\
\hline
in context &  45.00 (9/20) & 65.00 (13/20)  \\
w/o context & 25.00 (5/20) & 35.00 (7/20)  \\
\hline
\end{tabular}
\caption{\label{Tab:context}
GPT-4 performance with/without context.
}
\end{table}

\subsection{In-Context Learning}
In this section, we test the in-context learning ability of GPT-4. We observe that GPT-4 is prone to output more correct answers within the same conversation window after approximately eight conversation rounds; GPT-4's accuracy increases after seeing more examples. During this procedure, no feedback is provided to the conversation. 

To exemplify this phenomenon, we conduct an experiment on the LogiQA 2.0 ood data and the ConTRoL dataset, each representing a typical test case for the task of multi-choice reading comprehension and natural language inference. We randomly select 20 instances from each dataset for the following testing. Remember that GPT-4's performance is not competitive on these two datasets. 

We first test the 20 instances from the same conversation window; then, we test each instance of 20 in a new conversation window. The testing results are shown in Table \ref{Tab:context}
For the LogiQA 2.0 ood dataset, GPT-4 yields 9 correct answers when the 20 examples are in the same conversation window. However, without the context, the number of correct answers drops to 5. For the ConTRoL dataset, we find that GPT-4 answers 13 questions correctly with the context, and it drops to 7 without the context. 
Appendix \ref{sec:appendix_c} shows an example where GPT-4 answers the question correctly inside the context while does not give the correct answer in a new conversation window.

\subsection{Chain-of-Thought Prompting}
Chain-of-Thought (CoT) prompting is explored by many researchers and shows promising results on complex multi-step reasoning tasks \cite{kojima2023large}. 
This section explores zero-shot CoT prompting for GPT-4 on logical reasoning datasets. The experiment is conducted on the LogiQA 2.0 ood data. We choose the same 112 instances as we do manual tests with GPT-4, which is shown in Table~\ref{Tab:mcrq}. For this round, we add the prompt "Let's think step by step" to the instruction. By adding this prompt, GPT-4 generates longer reasoning texts exemplifying the reasoning steps. We collect the final answer for each CoT reasoning process and get 61 correct answers out of 112 questions, which is higher than the previous experiment without CoT prompting. 

Overall, the evaluation of the logical reasoning ability of ChatGPT and GPT-4 highlights the importance of developing more sophisticated benchmarks in textual inference to improve NLU systems' logical reasoning abilities further. The results also suggest that there is still room for improvement in language models' logical reasoning abilities, particularly when handling out-of-distribution datasets. Researchers need to continue developing more sophisticated benchmarks in textual inference to improve NLU systems' logical reasoning abilities further. Exploring new approaches to training language models that can better handle out-of-distribution datasets and other challenges associated with real-world applications is important.

\section{Conclusion}
We evaluate the logical reasoning ability of ChatGPT and GPT-4 on multiple logical reasoning datasets, including multi-choice reading comprehension and natural language inference tasks. We test on popular benchmarks, as well as less-known NLI datasets and newly-released ones. The experiment results show that ChatGPT and GPT-4 outperform the traditional fine-tuning method on most logical reasoning benchmarks, indicating that these two models can better reason logically. However, both models face challenges in handling new and out-of-distribution data. Besides, ChatGPT and GPT-4 do not perform astonishingly well on the natural language inference task requiring logical reasoning.

% Entries for the entire Anthology, followed by custom entries
\bibliography{anthology,custom}
\bibliographystyle{acl_natbib}

\appendix
\section{Instruction and Prompt for the multi-choice reading comprehension task}
\label{sec:appendix_a}

\texttt{
Instructions: You will be presented with a passage and a question about that passage. There are four options to be chosen from, you need to choose the only correct option to answer that question. If the first option is right, you generate the answer 'A', if the second option is right, you generate the answer 'B', if the third option is right, you generate the answer 'C', if the fourth option is right, you generate the answer 'D'. Read the question and options thoroughly and select the correct answer from the four answer labels. Read the passage thoroughly to ensure you know what the passage entails.
}

\section{Instruction and Prompt for the natural language inference task}
\label{sec:appendix_b}
\texttt{
Instructions: You will be presented with a premise and a hypothesis about that premise. You need to decide whether the hypothesis is entailed by the premise by choosing one of the following answers: 'e': The hypothesis follows logically from the information contained in the premise. 'c': The hypothesis is logically false from the information contained in the premise. 'n': It is not possible to determine whether the hypothesis is true or false without further information. Read the passage of information thoroughly and select the correct answer from the three answer labels. Read the premise thoroughly to ensure you know what the premise entails.}

\section{GPT-4 Answers with/without Context}
\label{sec:appendix_c}
Example:

The gold label is "B"

Question: 

Crimes involving the use of guns are more likely to lead to homicides than other types of crimes, but most crimes involving the use of guns do not lead to homicides. Therefore, it is not necessary to distinguish the illegal use of guns as a serious criminal offense from other criminal offenses in criminal law.

According to the logical loopholes in the above argument, which of the following options is the most similar?

A. Obese people are more likely to suffer from heart disease than normal-weight people. However, obesity accounts for only a small proportion of China's population. Therefore, in China, there is no need for medical and health circles to emphasize the risk of heart disease

B. Promiscuous sex is more likely to be infected with AIDS than regular sex. However, only a small proportion of people who have promiscuous sex are infected with AIDS. Therefore, there is no need to emphasize the harm of promiscuous sex in the publicity of AIDS prevention

C. The popular view is that smoking is more likely to cause lung cancer than non-smoking. However, in some countries, the proportion of lung cancer patients with a smoking history is not higher than that of the total population. Therefore, the above popular view is likely to be a prejudice

D. High-income people are more able to enjoy life than low-income people. But many high-income people claim to be unhappy. Therefore, the pursuers of a happy life do not have to pay attention to the level of income

Answer and reason:

GPT-4 output:

\textbf{With context:}

A. Obese people are more likely to suffer from heart disease than normal weight people. However, obesity accounts for only a small proportion of China's population. Therefore, in China, there is no need for medical and health circles to emphasize the risk of heart disease.

Reason: Both the original argument and option A have the same logical loophole. In the original argument, it is stated that crimes involving guns are more likely to lead to homicides but most don't, concluding that there is no need to distinguish the illegal use of guns as a serious criminal offense. Similarly, in option A, it is stated that obese people are more likely to suffer from heart disease but obesity accounts for only a small proportion of the population, concluding that there is no need to emphasize the risk of heart disease in China. Both arguments downplay the importance of addressing a specific risk or issue based on the fact that it doesn't affect a majority or large proportion of the population, even though the consequences are more severe for the affected individuals.

\textbf{Without context:}

B. Promiscuous sex is more likely to be infected with AIDS than regular sex. However, only a small proportion of people who have promiscuous sex are infected with AIDS. Therefore, there is no need to emphasize the harm of promiscuous sex in the publicity of AIDS prevention

Reason: Both the original argument and option B exhibit a similar logical flaw. The original argument acknowledges that gun-related crimes are more likely to lead to homicides, yet concludes that there is no need to treat such crimes as distinctively serious because most gun-related crimes do not lead to homicides. Similarly, option B acknowledges that promiscuous sex is more likely to result in AIDS infection, but concludes that there is no need to emphasize the harm of promiscuous sex in AIDS prevention efforts because only a small proportion of people engaging in promiscuous sex contract AIDS. Both arguments downplay the increased risk associated with specific behaviors or situations by focusing on the fact that not all instances of those behaviors or situations lead to negative outcomes.

\end{document}